\documentclass{article}
\usepackage{ijcai26}

\usepackage{times}
\usepackage{soul}
\usepackage{url}
\usepackage[hidelinks]{hyperref}
\usepackage[small]{caption}
\usepackage{graphicx}
\usepackage{amsmath}
\usepackage{amsthm}
\usepackage{booktabs}
\usepackage{algorithm}
\usepackage{algorithmic}
\usepackage{graphicx}
\usepackage{acronym}
\usepackage{booktabs}
\usepackage{caption}
\usepackage{multirow}
\usepackage[table]{xcolor}
\usepackage{array}
\usepackage{subcaption}
\usepackage{hyperref}
\usepackage{pifont}
\definecolor{graybg}{gray}{0.95}
\definecolor{cgreen}{RGB}{0, 180, 0}
\definecolor{cred}{RGB}{200, 0, 0}

\newcommand{\yes}{\textcolor{cgreen}{\ding{51}}} 
\newcommand{\no}{\textcolor{cred}{\ding{55}}}

\usepackage[scheme=plain,fontset=none]{ctex}
\usepackage{fontspec}
\setCJKsansfont[
BoldFont=FandolHei-Bold.otf,
ItalicFont=FandolHei-Regular.otf
]{FandolHei-Regular.otf}

\setCJKmonofont[
BoldFont=FandolFang-Regular.otf,
ItalicFont=FandolFang-Regular.otf
]{FandolFang-Regular.otf}
\usepackage[most]{tcolorbox}
\tcbuselibrary{theorems}
\tcbuselibrary{breakable}
\usepackage{amssymb}

\newtcbtheorem[auto counter]{prm}{Prompt}{
  colbacktitle = blue!60!white,  
  colframe = blue!60!white,     
  colback = blue!5!white,       
  breakable,
  enhanced jigsaw,
  sharp corners,
  fontupper = \rmfamily\small,
  arc = 2pt,
  boxsep = 5pt,             
  left = 2pt,            
  right = 2pt,              
  top = 2pt,          
  bottom = 2pt,             
  title filled = true,
}{t}
\newtcbtheorem[auto counter]{ch}{提示词}{
  colbacktitle = blue!60!white,
  colframe = blue!60!white,
  colback = blue!5!white,
  breakable,
  enhanced,
  sharp corners,
  fonttitle=\bfseries,
  fontupper = \rmfamily\small,
  arc = 2pt,
  boxsep = 5pt,
  left = 5pt,
  right = 5pt,
  top = 5pt,
  bottom = 5pt,
  title filled = true,
}{t}

\newtcbtheorem[auto counter]{case}{Case}{
  colbacktitle = orange!60!white,  
  colframe = orange!60!white,     
  colback = orange!5!white,       
  breakable,
  enhanced jigsaw,
  sharp corners,
  fontupper = \rmfamily\small,
  arc = 2pt,
  boxsep = 5pt,             
  left = 2pt,            
  right = 2pt,              
  top = 2pt,          
  bottom = 2pt,             
  title filled = true,
}{t}

\acrodef{LLMs}{Large Language Models}
\acrodef{SFT}{Supervised Fine-Tuning}
\acrodef{AI4SS}{AI for Social Science}
\acrodef{AI}{artificial intelligence}
\acrodef{CJ}{Critique-and-Judge}
\acrodef{HATD}{Hierarchical Atomic Task Decomposition}
\acrodef{Q-A}{Question-Answer pairs}

\hypersetup{
  pdfinfo={
    TemplateVersion={IJCAI.2026.0}
  }
}

\title{EduResearchBench: A Hierarchical Atomic Task Decomposition Benchmark for Full-Lifecycle Educational Research}

\author{
	Houping~Yue$^{1,\dagger}$\and
	Zixiang~Di$^{3\dagger}$\and
	Mei~Jiang$^1$\and 
	Bingdong~Li$^{1,*}$\and \\
	Hao~Hao$^1$\and
	Yu~Song$^1$\and
	Bo~Jiang$^1$\and
	Aimin~Zhou$^{2,1}$\\
	\affiliations
	$^1$Shanghai Institute of AI for Education, East China Normal University\\
	$^2$Shanghai Innovation Institute\\
	$^3$School of Computer Science and Technology, East China Normal University\\
	\emails
	yuehouping@gmail.com,
	zixiang1119@163.com,
	51265901071@stu.ecnu.edu.cn,
	bdli@cs.ecnu.edu.cn,\\
	\{hhao, ysong\}@mail.ecnu.edu.cn,
	bjiang@deit.ecnu.edu.cn,
	amzhou@cs.ecnu.edu.cn
}

\begin{document}

\maketitle
\renewcommand{\thefootnote}{\fnsymbol{footnote}} 
\footnotetext[2]{These authors contributed equally to this work.}
\footnotetext[1]{Corresponding author.}
\renewcommand{\thefootnote}{\arabic{footnote}}

\begin{abstract}
While Large Language Models (LLMs) are reshaping the paradigm of AI for Social Science (AI4SS), rigorously evaluating their capabilities in scholarly writing remains a major challenge. Existing benchmarks largely emphasize single-shot, monolithic generation and thus lack the fine-grained assessments required to reflect complex academic research workflows. To fill this gap, we introduce EduResearchBench, the first comprehensive evaluation platform dedicated to educational academic writing. EduResearchBench is built upon our Hierarchical Atomic Task Decomposition (HATD) framework, which decomposes an end-to-end research workflow into six specialized research modules (e.g., Quantitative Analysis, Qualitative Research, and Policy Research) spanning 24 fine-grained atomic tasks. This taxonomy enables an automated evaluation pipeline that mitigates a key limitation of holistic scoring, where aggregate scores often obscure specific capability bottlenecks, and instead provides fine-grained, diagnostic feedback on concrete deficiencies. Moreover, recognizing the high cognitive load inherent in scholarly writing, we propose a curriculum learning strategy that progressively builds competence from foundational skills to complex methodological reasoning and argumentation. Leveraging 55K raw academic samples, we curate 11K high-quality instruction pairs to train EduWrite, a specialized educational scholarly writing model. Experiments show that EduWrite (30B) substantially outperforms larger general-purpose models (72B) on multiple core metrics, demonstrating that in vertical domains, data quality density and hierarchically staged training curricula are more decisive than parameter scale.
\end{abstract}

\section{Introduction \label{intro}}
\ac{LLMs} reshape the \ac{AI4SS} paradigm, and the utilization of \ac{LLMs} to assist academic research serves as a prevalent practice~\cite{wang2023scientific,tang2025ai,yu2025survey,an2024make}. 
This shift empowers researchers to leverage computational intelligence for simulating complex social dynamics and analyzing qualitative data at scale~\cite{xu2024ai}. 
Particularly within the domain of education, the application of LLMs evolves from simple instructional \ac{Q-A} to rigorous scholarly production~\cite{sun2024scieval,xu2025edubench}. 
In the realm of academic writing, researchers investigate the potential of \ac{LLMs} to transcend basic text generation, functioning instead as autonomous research assistants capable of executing higher-order cognitive tasks such as formulating hypotheses, structuring arguments, and synthesizing literature reviews. 
However, educational academic writing is characterized by high cognitive load, stemming from the simultaneous requirements of heterogeneous information synthesis, long-horizon logical consistency, and strict adherence to academic norms~\cite{lazer2009life,bail2024can}.

Despite these advancements, evaluating the proficiency of \ac{LLMs} in such complex workflows remains a significant challenge. Existing benchmarks primarily exhibit three fundamental limitations: 
1) Current benchmarks primarily focus on general instruction following or long-text generation tasks (e.g., LongWriter~\cite{bailongwriter}), often prioritizing linguistic fluency and surface plausibility. However, educational academic research is characterized by high cognitive load, requiring models to possess profound theoretical depth and logical consistency. Existing evaluation systems struggle to capture the academic rigor and adherence to norms in model-generated content.
2) Existing evaluations predominantly frame the task as end-to-end text generation, collapsing the complex research process into a single output. This formulation breaks the inherent reasoning chain of academic research and fails to cover the full lifecycle, from topic ideation and literature-review synthesis to methodological argumentation and final synthesis.
3) Traditional evaluations typically rely on holistic scoring, providing only a coarse-grained score. This approach fails to distinguish whether low scores stem from a lack of domain knowledge or a deficit in specific atomic capabilities (e.g., citation norms or variable operationalization).

To address the aforementioned challenges, we propose the \ac{HATD} framework, which comprises the following:
1) We propose a Task-Level Curriculum Learning strategy. Instead of demanding immediate end-to-end generation, this strategy stages the training process from foundational capabilities (e.g., topic recommendation) to complex methodological reasoning (e.g., literature synthesis).
2) We decompose the complex research workflow into six specialized research modules (e.g., Quantitative Research, Qualitative Research) and 24 fine-grained atomic tasks, enabling precise modeling of the full research lifecycle.
3)Based on this taxonomy, we construct an automated evaluation pipeline. This pipeline overcomes the ambiguity of traditional holistic scoring, providing precise diagnostic granularity for specific atomic capability deficits.
To empirically validate the effectiveness of this framework, we curate 11k high-quality instruction pairs from over 55k raw academic samples to train EduWrite via \ac{SFT}, a specialized model dedicated to scholarly writing~\cite{ouyang2022training}.

To summarize, our main contributions are as follows:
\begin{itemize}
    \item We construct EduResearchBench, filling the gap in evaluating full-lifecycle educational academic writing. It transcends the limitations of fragmented tasks inherent in existing benchmarks, covering the complete research workflow from topic ideation and literature review to results analysis, thereby achieving a systematic evaluation of high-cognitive-load writing tasks.
    \item We propose the \ac{HATD} framework and an automated evaluation pipeline. By decoupling complex writing into 24 atomic tasks, we facilitate a transition from holistic scoring to interpretable atomic capability evaluation, enabling the precise identification of specific deficits in domain knowledge and adherence to logical norms.
    \item We validate that EduWrite, trained via our Curriculum Learning strategy and high-quality instruction fine-tuning, significantly outperforms larger general-purpose models. This result demonstrates that in vertical domains, data density and hierarchical training curricula are more critical determinants of performance than mere parameter scale.
\end{itemize}

\begin{table*}[t]
    \centering
    \caption{Comparison of \textbf{AI4ResearchBench} with existing benchmarks. Our work is the only one covering the \textit{Full Lifecycle} of research with \textit{Atomic} granularity.}
    \label{tab:comparison}
    \resizebox{\textwidth}{!}{
    \begin{tabular}{l|c|cc|ccc|c}
        \toprule
        \multirow{2}{*}{\textbf{Benchmark}} & \multirow{2}{*}{\textbf{Lang.}} & \multicolumn{2}{c|}{\textbf{Evaluation Scope}} & \multicolumn{3}{c|}{\textbf{Task Granularity \& Features}} & \multirow{2}{*}{\textbf{Core Focus}} \\
        \cmidrule(lr){3-4} \cmidrule(lr){5-7}
        & & \textbf{Domain} & \textbf{Source} & \textbf{Atomic?} & \textbf{Lifecycle?} & \textbf{Long-Ctx?} & \\
        \midrule
        OmniEduBench & Zh & Education & K12/College & \no & \no & \no & Knowledge QA \\
        AcademicEval & En & Academic & arXiv & \no & \no & \yes & Section Gen. \\
        WritingBench & Zh/En & General & 6 Domains & \no & \no & \no & Writing Quality \\
        LongBench-Write & Zh/En & General & - & \no & \no & \yes & Length Limit \\
        \midrule
        \textbf{EduResearchBench} & \textbf{Zh/En} & \textbf{Education} & \textbf{Academic} & \yes & \yes & \yes & \textbf{EduResearch} \\
        \bottomrule
    \end{tabular}
    }
\end{table*}

\section{Benchmark \label{benchmark}}
EduResearchBench is an academic writing benchmark dedicated to the education domain. It simulates the complex lifecycle of educational academic research, extending core evaluation dimensions beyond linguistic fluency to encompass logical consistency, methodological rigor, and strict adherence to domain-specific academic norms.
The construction of EduResearchBench is grounded in the \ac{HATD} framework. This framework deconstructs the complex research workflow into six specialized research modules and 24 fine-grained atomic tasks. This hierarchical design supports a novel automated evaluation mechanism: model performance is no longer assessed via coarse-grained holistic scoring but is independently evaluated at the atomic task level.

\begin{figure*}[ht]
    \centering
    \includegraphics[width=15cm]{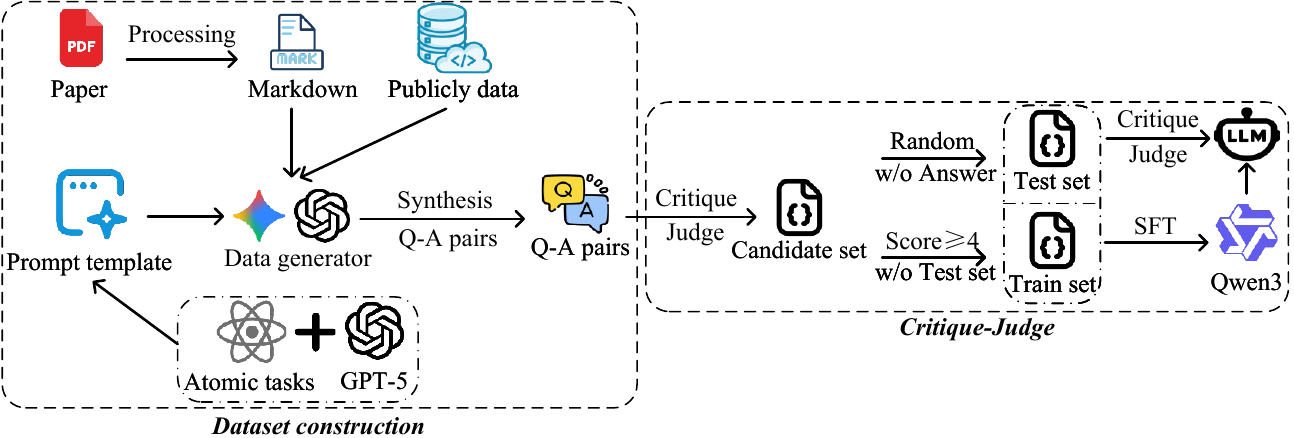} 
    \caption{\textbf{Overview of the EduResearchBench framework.} 
    GPT-5 and task templates are used to generate Q-A pairs from processed literature and public data. High-quality samples (score $\geq$ 4) are used to fine-tune Qwen3-30B-A3B, while a randomly selected subset serves as the test set for performance evaluation.}
    \label{fig1:overview}
\end{figure*}

\subsection{Task Definition}
To systematically evaluate the complex competencies required for educational research, we establish the \ac{HATD} framework. This taxonomy deconstructs the complete lifecycle of educational academic research into six specialized research modules, covering the entire workflow from topic ideation and drafting of research content to normative review.

\begin{figure*}[t]
    \centering
    \begin{minipage}[c]{0.53\textwidth}
        \centering
        \captionof{table}{The \ac{HATD} Taxonomy of EduResearchBench. It consists of 6 specialized research modules and 24 fine-grained atomic tasks.}
        \label{tab:taxonomy}
        \resizebox{\linewidth}{!}{%
            \begin{tabular}{@{}lp{7.5cm}@{}} 
            \toprule
            \textbf{Module} & \textbf{Atomic Tasks} \\ \midrule
            \multirow{4}{*}{\textbf{Topic Rec.}} & 1. Hotspot Extraction \\
             & 2. Trend Analysis \\
             & 3. Emerging Domain Exploration \\
             & 4. Specific Topic Recommendation \\ \midrule
            \multirow{5}{*}{\textbf{Quantitative}} & 1. Structure \& Hypotheses \\
             & 2. Variables \& Operationalization \\
             & 3. Sampling \& Procedures \\
             & 4. Reliability \& Validity \\
             & 5. Stats \& APA Format \\ \cmidrule(l){1-2}
            \multirow{4}{*}{\textbf{Qualitative}} & 1. Thick Description \\
             & 2. Analysis Process Write-up \\
             & 3. Findings Generation \\
             & 4. Trustworthiness Statement \\ \cmidrule(l){1-2}
            \multirow{4}{*}{\textbf{Policy Analysis}} & 1. Policy Structure Extraction \\
             & 2. Stakeholder Impact \\
             & 3. Term Interpretation \\
             & 4. Eval. \& Recommendation \\ \midrule
            \multirow{5}{*}{\textbf{Theory/Survey}} & 1. Argument Structure \\
             & 2. Core Concept Def. \\
             & 3. Theory Critique \\
             & 4. Integ. \& Innov. \\
             & 5. Contr. \& Edu. \\ \midrule
            \multirow{2}{*}{\textbf{Peer Review}} & 1. Language Quality \\
             & 2. Logic Stream \\ \bottomrule
            \end{tabular}%
        }
    \end{minipage}%
    \hfill
    \begin{minipage}[c]{0.45\textwidth}
        \centering
        \includegraphics[width=\linewidth]{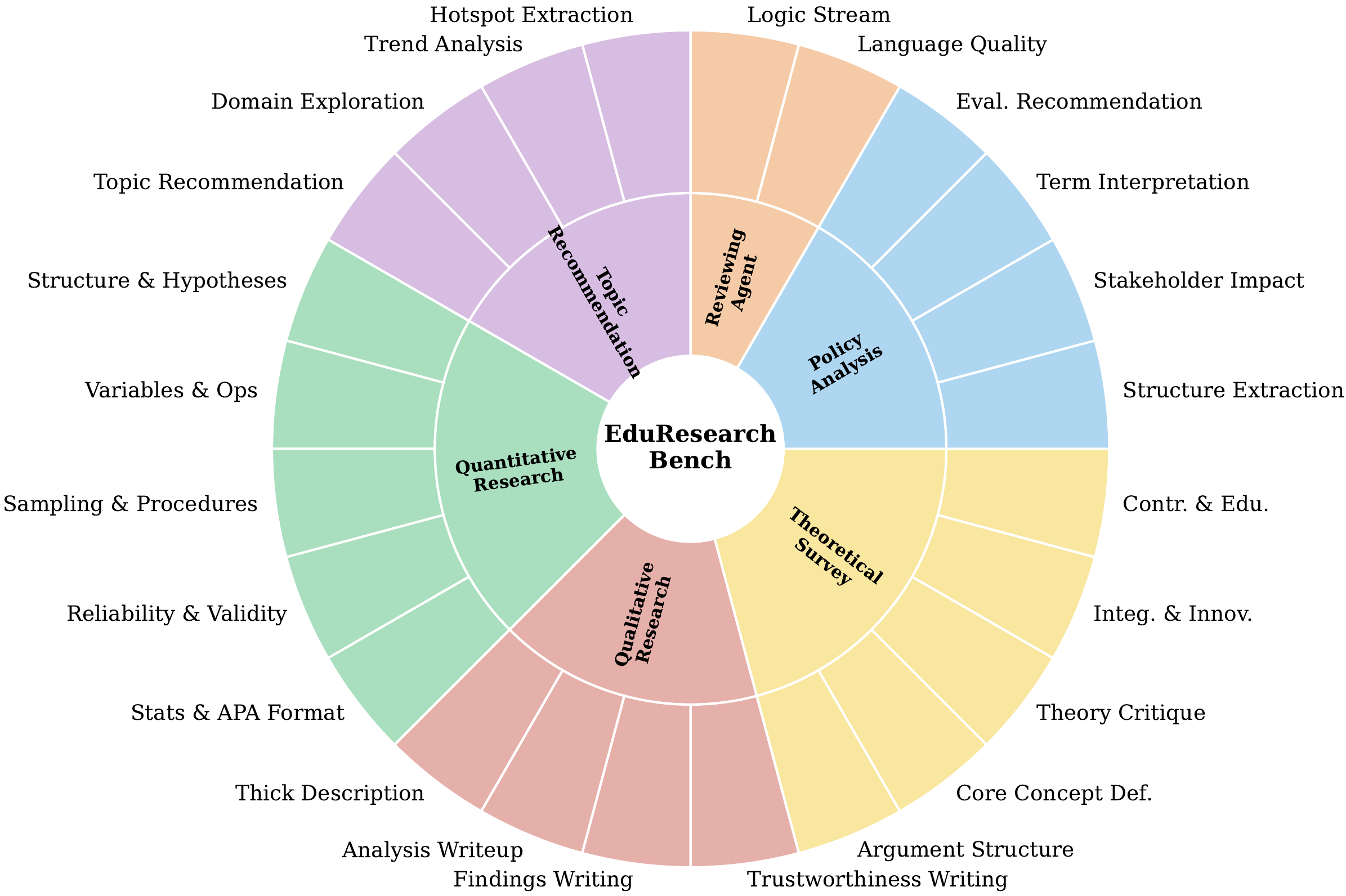} 
        \captionof{figure}{The inner ring represents specialized research modules, while the outer ring displays the distribution of fine-grained atomic tasks.}
        \label{fig:taxonomy_sunburst}
    \end{minipage}
\end{figure*}

As shown in Table~\ref{tab:taxonomy} and Figure~\ref{fig:taxonomy_sunburst}, these modules cover 24 fine-grained atomic tasks. Each task represents a distinct functional unit within the logical chain of academic writing, ranging from foundational information extraction (e.g., Hotspot Extraction) to complex methodological construction (e.g., Variables Operationalization or Theory Critique). This granular decomposition achieves precise evaluation of the model's performance throughout the educational academic research process.

\subsection{Benchmark Construction}
The construction of EduResearchBench follows a rigorous LLM-driven data development pipeline and is seamlessly integrated with the Supervised Fine-Tuning (SFT) workflow. As shown in Figure 2, this pipeline consists of three key stages: policy-driven data synthesis, dual-judge filtering, and standardized dataset partitioning.

Acquiring educational academic writing data characterized by deep theoretical grounding and complex logical structures presents a significant challenge. To address this, we leverage GPT-5 and Gemini3-Flash as generation models to synthesize a large-scale corpus of instruction–response pairs.
\subsubsection{Prompt Engineering and Data Construction Pipeline}
\textbf{Policy-Guided Prompt Engineering:} We design highly customized prompt templates for each of the 24 atomic tasks. Unlike automatically generated prompts, these templates are crafted by human domain experts and undergo multiple rounds of manual iterative refinement to ensure the generated content meets expert-level academic standards.

\noindent \textbf{Data Format:} The synthesized data is formatted as JSONL (in Messages format) to support direct \ac{SFT} implementation. Each data entry is assigned a unique ID for tracking and matching. This format aligns with mainstream \ac{SFT} datasets, ensuring compatibility with widely-used training frameworks.
\subsubsection{Dual-Model Critique-and-Judge Filtering Mechanism}
To ensure the reliability of evaluation results and mitigate the inherent systematic biases of using LLMs as judges, we implement a dual-judge verification mechanism.
We select Doubao-seed 1.6 and DeepSeek-V3.2 as evaluation judges. These models are chosen for their exceptional reasoning capabilities and strong adherence to complex instructions.

\noindent \textbf{Two-Stage Evaluation Process:} For each synthesized Q-A pair, both judge models execute the following two-step evaluation:
\begin{itemize}
    \item \textbf{Critical Analysis:} The judge model first generates a qualitative critique, analyzing the answer's logical coherence, theoretical depth, and adherence to the policy constraints embedded in the prompt.
    \item \textbf{Quantitative Scoring:} Based on the critical analysis, the model provides a quantitative score on a scale of 1 to 5 within the same response cycle.
\end{itemize}
\subsubsection{Dataset Partitioning and Scoring Strategy}
Based on the evaluation results from the dual-model judges, we apply strict data filtering and partitioning strategies to construct the training and test sets.

\noindent \textbf{SFT Training Set Construction:} We adopt a "High-Standard" criterion. Only data entries achieving a comprehensive score of $\ge$ 4.0 from the dual judges are included in the SFT training set. This ensures that the training data serves as high-quality demonstrations, effectively aligning baseline models with educational research standards.

\noindent \textbf{Test Set Construction:} To construct an unbiased benchmark, we employ a stratified sampling strategy to independently partition a test set from the corpus, processed into two versions:
\begin{itemize}
    \item \textbf{Reference Set:} Contains the original questions and their corresponding high-quality reference answers (Gold Answers), used for benchmarking against top-tier closed-source models (e.g., GPT-5).
    \item \textbf{Inference Set:} With the "Answer" content removed (label-stripped) to prevent data leakage, serving as instruction inputs to evaluate the generation capabilities of other models.
\end{itemize}

\section{Experiments \label{experimrnts}}
\subsection{Experimental Setup}
\subsubsection{Datasets}
EduResearchBench is constructed under our \ac{HATD} taxonomy, which decomposes the educational research workflow into 6 specialized research modules and 24 fine-grained atomic tasks. The benchmark follows an LLM-driven data construction pipeline and is seamlessly integrated with \ac{SFT}.

As summarized in Table~\ref{tab:final_stats}, we start from 55,493 raw academic samples and curate 11,357 high-quality \ac{Q-A} for \ac{SFT} training, together with a held-out 5,300 test set. The token statistics in Table~\ref{tab:final_stats} highlight the long-context nature of our training data: the average sequence length reaches 3,136 tokens, and the maximum length exceeds 100K tokens (108,520 tokens), which better reflects realistic educational scholarly writing scenarios.

For evaluation, we construct two test-set variants following Section~2.2.3: (i) a \textbf{Standard} split that retains reference answers for reference-based analysis (e.g., benchmarking against frontier systems), and (ii) a \textbf{Reasoning} split where reference answers are removed to prevent leakage and to evaluate models purely based on generated outputs. Unless otherwise specified, we report main results on the Reasoning split and use the Standard split for reference-based diagnostics.
\begin{table}[t]
    \centering
    \small
    \renewcommand{\arraystretch}{1.3}
    \caption{Statistics of \textbf{EduRessearchBench}. The table reports the data scale across different splits and the token distribution of the training set, highlighting the long-context characteristics. Token statistics are computed using the Qwen series tokenizer (Qwen/Qwen2.5-7B-Instruct).}
    \label{tab:final_stats}
    \setlength{\tabcolsep}{18pt} 
    \begin{tabular}{lr}
        \toprule
        \textbf{Category} & \textbf{Num} \\
        \midrule
        \rowcolor{graybg} \multicolumn{2}{l}{\textit{\textbf{Data Scale}}} \\
        Raw Data & 55,493 \\
        Training Set (SFT) & 11,357 \\
        Test Set & 5,300 \\
        \midrule 
        \rowcolor{graybg} \multicolumn{2}{l}{\textit{\textbf{Token Statistics (Training Set)}}} \\
        Total Tokens & $\approx$ 35.6M \\
        Average Tokens per Sample & 3,136 \\
        Max. Tokens per Sample & 108,520 \\
        Min. Tokens per Sample & 472 \\
        
        \bottomrule
    \end{tabular}
\end{table}

\subsubsection{Evaluation Protocol}
\paragraph{Models.}
We evaluate both proprietary and open-source \ac{LLMs} with varying parameter scales. The closed-source models include GPT-5 and Gemini 3 Flash. For open-source models, we focus on Qwen-2.5-72B-Instruct~\cite{yang2024qwen2} and Llama-3-70B-Instruct~\cite{grattafiori2024llama}. Additionally, we assess capability-enhanced LLMs, which include WritingBench (Qwen2.5-7B)~\cite{wu2025writingbench}, LongWriter (GLM4-9B)~\cite{bailongwriter}, InnoSpark-72B, and EduWrite, a model fine-tuned via \ac{SFT} on the Qwen3-30B-A3B~\cite{yang2025qwen3}. This classification allows for a clear comparison across different model types, highlighting variations in model capabilities and performance.

\paragraph{Inference.}
For each test instance, the evaluated model generates a single response conditioned on the task-specific prompt. To ensure fair comparison, we keep decoding configurations fixed across all evaluated open-source models and report the full inference hyperparameters (e.g., max new tokens, top-$p$, temperature) in the appendix. We deploy open-source models with vLLM on $8\times$ NVIDIA H200 GPUs using tensor parallelism (TP=8).

\paragraph{Automated scoring.}
We adopt an automated evaluation pipeline consistent with our benchmark construction. Each model output is independently assessed by two judges, Doubao-seed-1.6~\cite{guo2025seed1} and DeepSeek-V3.2~\cite{liu2025deepseek}, using a two-stage procedure: the judge first produces a qualitative critique that examines logical soundness, theoretical depth, and adherence to task constraints, and then assigns a scalar score on a 1-5 scale. The final instance score is computed as the 50/50 average of the two judges.
We report results at three granularities: overall performance, module-level aggregates (6 modules), and atomic-task-level scores (24 tasks), enabling fine-grained diagnosis of 
capability bottlenecks.

\subsubsection{Model Training}
\paragraph{Supervised fine-tuning (EduWrite).}
To validate whether benchmark-driven, high-quality supervision improves educational scholarly writing, we perform full-parameter \ac{SFT} on Qwen3-30B-A3B using the Ms-Swift framework, producing our domain-specialized model EduWrite. The SFT training set is curated via Dual-Judge Filtering: a sample is included only if it receives scores $\ge 4.0$ from both judges, ensuring high-quality demonstrations aligned with expert-level academic standards.

\begin{figure}[htbp]
\centering
\includegraphics[width=0.48\textwidth]{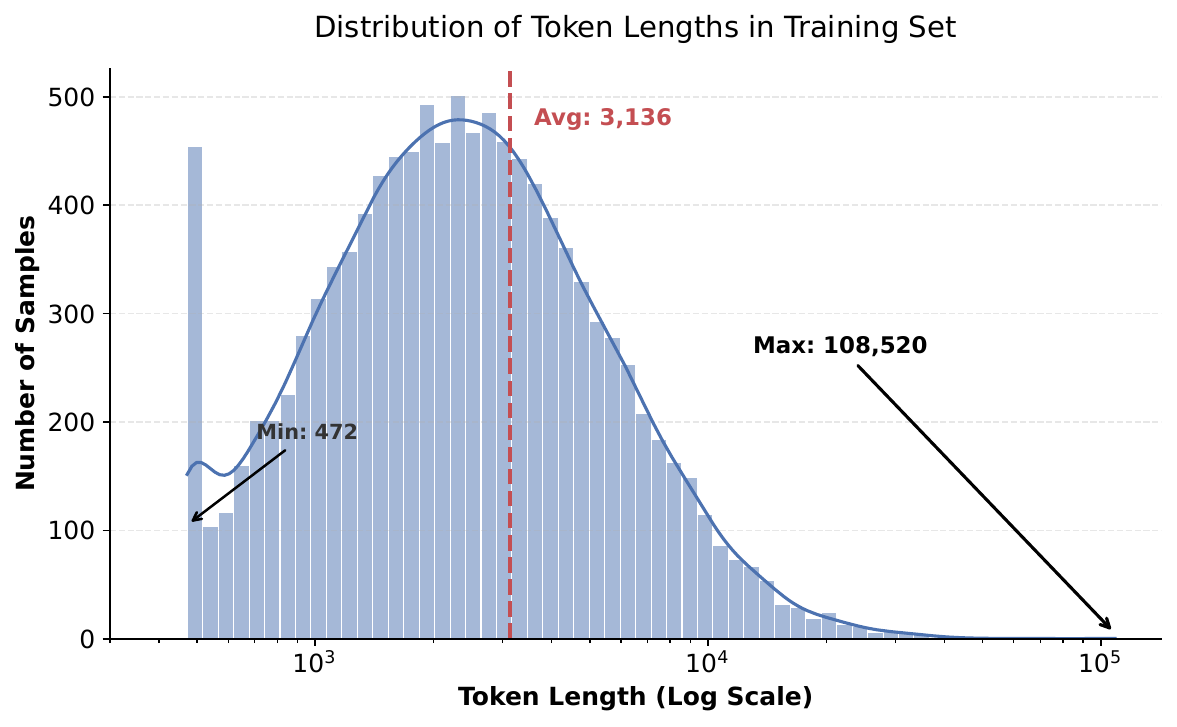}
\caption{Distribution of Token Lengths in Training Set. The histogram shows the distribution of token lengths (on a logarithmic scale) with the minimum token length of 472, the average of 3,136, and the maximum of 108,520.}
\label{fig:token_distribution}
\end{figure}

\begin{table*}[t]
\centering
\small
\renewcommand{\arraystretch}{1.2}
\setlength{\tabcolsep}{8pt}
\caption{Main experimental results across different agents.}
\label{tab:main_results}
\begin{tabular}{lccccccc}
\toprule
\textbf{Model} & \textbf{Overall} & \textbf{Reviewer} & \textbf{Quant.} & \textbf{Qual.} & \textbf{Topic} & \textbf{Policy} & \textbf{Survey} \\
\midrule

\rowcolor{gray!10} \multicolumn{8}{l}{\textit{Closed-source Models}} \\
GPT-5 &
\textbf{3.63} & \underline{4.03} & \textbf{3.12} & \textbf{3.65} & \textbf{3.96} & \textbf{3.58} & \textbf{3.63} \\
Gemini 3 Flash &
\underline{3.34} & \textbf{4.07} & \underline{2.92} & \underline{3.45} & \underline{3.82} & \underline{3.36} & \underline{3.21} \\
\midrule

\rowcolor{gray!10} \multicolumn{8}{l}{\textit{Open-source LLMs}} \\
Qwen-2.5-72B-Instruct & 2.90 & 3.41 & 2.72 & 3.01 & 3.28 & 2.62 & 2.42 \\
Llama-3-70B-Instruct & 2.46 & 2.50 & 2.51 & 2.67 & 2.81 & 2.34 & 1.50 \\
\midrule

\rowcolor{gray!10} \multicolumn{8}{l}{\textit{Capability-enhanced LLMs}} \\
WritingBench (Qwen2.5-7B) & 2.44 & 2.40 & 2.19 & 2.33 & 3.00 & 2.62 & 1.71 \\
LongWriter (GLM4-9B) & 2.34 & 2.62 & 2.27 & 2.55 & 2.43 & 2.26 & 1.77 \\
InnoSpark-72B & 2.97 & 3.31 & 2.75 & 3.06 & 3.17 & 2.98 & 2.56 \\
EduWrite (30B) & 3.20 & 3.33 & 2.87 & 3.07 & 3.72 & 3.32 & 2.74 \\
\bottomrule
\end{tabular}
\end{table*}

\begin{figure}[htbp]
\centering
\includegraphics[width=0.48\textwidth]{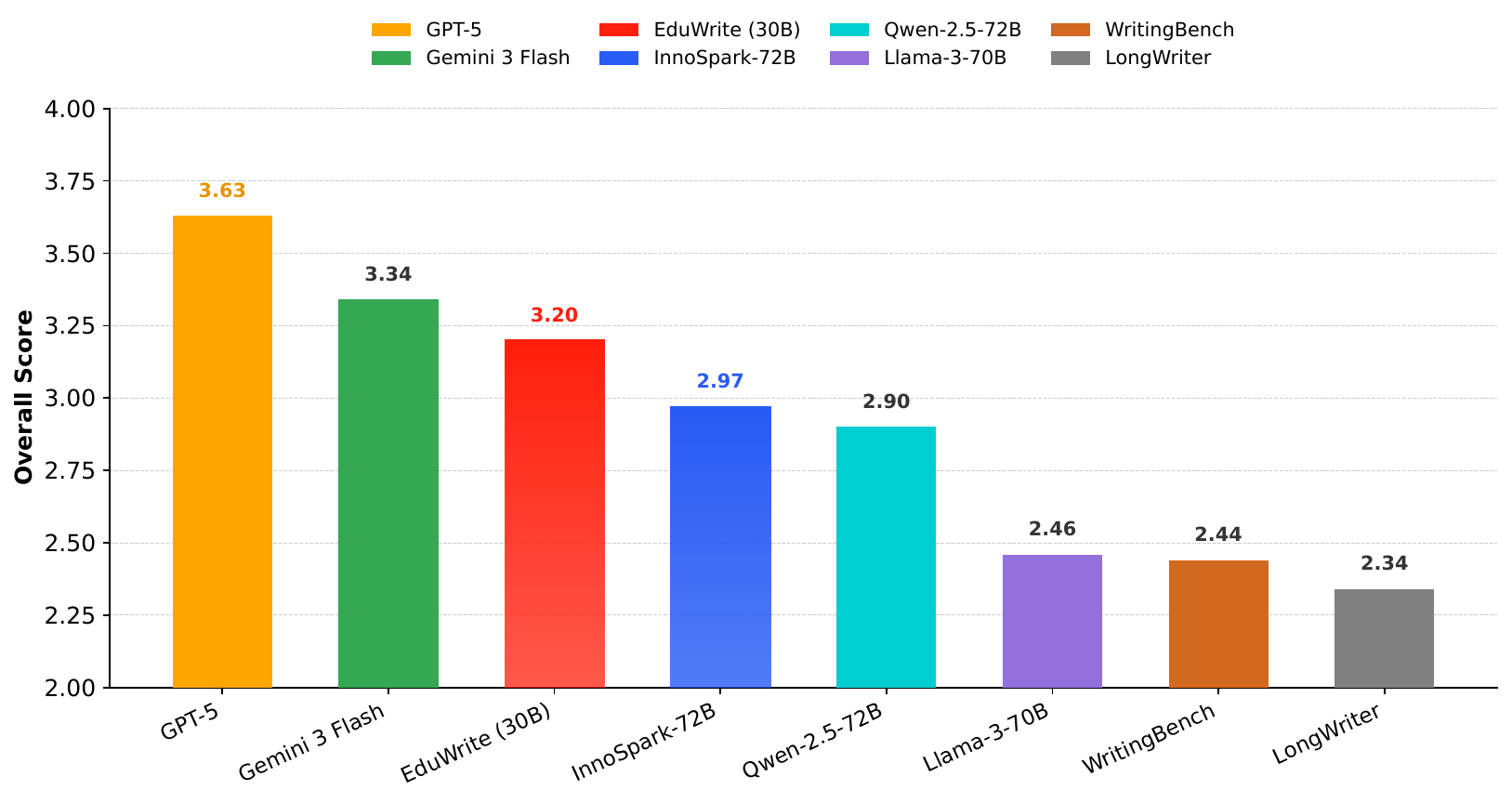}
\caption{Overall performance of different models. The models are evaluated based on their overall scores, showing the comparison between closed-source models, open-source models, and capability-enhanced LLMs.}
\label{fig:overall_scores}
\end{figure}

\subsection{Analysis and Findings}
Table~\ref{tab:main_results} presents the performance results of different \ac{LLMs} on EduResearchBench, while Figure~\ref{fig:overall_scores} visualizes the overall performance comparison of these models.

\begin{table*}[t]
\centering
\small
\renewcommand{\arraystretch}{1.15}
\setlength{\tabcolsep}{7pt}
\caption{\textbf{Curriculum learning ablation} on EduResearchBench. 
\textbf{EduWrite\_SFT} denotes one-stage SFT on the full training set. 
\textbf{Stage-1} and \textbf{Stage-2} denote curriculum-based training where Stage-2 continues from Stage-1.
All scores are obtained using \textbf{Doubao-seed-1.6} as the judge (1-5 scale). 
\textbf{Bold} indicates the best result in each row.}
\label{tab:curriculum_learning}

\begin{tabular}{lccc}
\toprule
\textbf{Metric (Mean)} & \textbf{EduWrite\_SFT} & \textbf{EduWrite\_Stage-1} & \textbf{EduWrite\_Stage-2} \\
\midrule
Overall                 & 3.63 & 3.58 & \textbf{3.81} \\
Reviewer (avg.)         & 3.48 & \textbf{3.72} & 3.65 \\
Quantitative (avg.)     & 3.07 & 2.98 & \textbf{3.08} \\
Qualitative (avg.)      & 4.05 & 3.79 & \textbf{4.06} \\
Topic (avg.)            & 4.12 & 4.12 & \textbf{4.19} \\
Policy (avg.)           & 3.49 & 3.53 & \textbf{4.19} \\
\bottomrule
\end{tabular}

\vspace{3pt}
\footnotesize
\textit{Note.} \textbf{Bold} indicates the best performance and \underline{underlined} indicates the second-best for each column.

\end{table*}

\subsection{Curriculum Learning Ablation}
To examine whether curriculum learning benefits educational scholarly writing, we compare one-stage SFT with a staged curriculum strategy. 
EduWrite\_SFT performs standard one-stage SFT on the full curated training set. 
In contrast, EduWrite\_Stage-1 and EduWrite\_Stage-2 implement a curriculum schedule, where Stage-2 continues training from the Stage-1 checkpoint with a more advanced (or broader) task mixture.

Table~\ref{tab:curriculum_learning} summarizes the results.\footnote{For this ablation, we use Doubao-seed-1.6 as a single judge to reduce evaluation cost; we observe consistent trends with the full evaluation protocol.} 
Overall, the staged curriculum yields clear gains: EduWrite\_Stage-2 achieves the best overall score (3.81), improving over one-stage SFT (3.63). 
The improvement is particularly pronounced in \textit{Policy Analysis} (4.19 vs.\ 3.49), suggesting that curriculum-based training better supports higher-cognitive-load modules that require structured reasoning and domain-grounded argumentation. 
We also observe consistent but smaller gains on \textit{Topic Recommendation} (4.19 vs.\ 4.12) and \textit{Qualitative Research} (4.06 vs.\ 4.05), while \textit{Quantitative Research} remains a bottleneck with only marginal changes (3.08 vs.\ 3.07). 
Interestingly, Stage-1 alone improves the \textit{Reviewer} module (3.72), whereas Stage-2 slightly trades off reviewer performance for larger gains in policy-oriented reasoning, indicating that later-stage training shifts capacity toward more complex, domain-specific writing skills.

\subsection{Human Expert Evaluation}
These results support our hypothesis that curriculum learning provides a practical training recipe for progressively building scholarly writing competence from foundational abilities to more demanding methodological and policy reasoning tasks.

To assess the reliability of EduResearchBench and the alignment between automated evaluation and human judgment, we conduct an expert human evaluation. We select 100 independent samples from the EduResearchBench test set, ensuring full coverage of the 24 atomic tasks.  
We compare the outputs of GPT-5 and Gemini 3 Flash. Five experienced human experts evaluate the responses based on three dimensions: content quality, logical consistency, and adherence to academic standards. The experts compare the model outputs for each sample.  
The evaluation is conducted in a pairwise comparison format, with randomized sample order to minimize bias. Expert ratings are recorded as A (better), B (better), or Tie.  
This evaluation provides insights into the consistency between human and automated assessments and offers guidance for refining the automated evaluation system.

\paragraph{Finding 1: EduResearchBench reveals stage-wise capability differences across the scholarly research lifecycle.}
Closed-source frontier models outperform capability-enhanced open-source models and achieve the highest overall scores (GPT-5: 3.63; Gemini 3 Flash: 3.34). However, the performance gaps are uneven across research modules, indicating that scholarly writing competence is not a monolithic skill. For instance, both closed-source models excel in the Reviewer module, with GPT-5 reaching 4.03 and Gemini 3 Flash achieving 4.07, while their performance in the Quantitative Research module is much lower (GPT-5: 3.12; Gemini 3 Flash: 2.92). This pattern supports our motivation: end-to-end holistic evaluation obscures where models truly succeed or fail, while modular evaluation exposes capability bottlenecks along the research lifecycle.

\paragraph{Finding 2: EduWrite leads open-source models.}
Among all capability-enhanced models, EduWrite (30B) achieves the best overall performance (3.20), outperforming InnoSpark-72B (2.97) and other writing-oriented baselines such as WritingBench (Qwen2.5-7B) (2.44) and LongWriter (GLM4-9B) (2.34). EduWrite's advantage is particularly pronounced in the Topic Recommendation (3.72) and Policy Analysis (3.32) modules, where it performs significantly better than other models. This result indicates that domain-specialized supervised fine-tuning on high-quality, long-context instruction data effectively transfers beyond generic long-form generation to education-specific research modules.

\paragraph{Finding 3: Parameter scale alone does not guarantee stronger educational scholarly writing.}
A notable result is that EduWrite (30B) surpasses the much larger open-source model InnoSpark-72B in overall performance (3.20 vs. 2.97). This advantage is primarily driven by education-relevant modules such as Topic Recommendation (3.72 vs. 3.17) and Policy Analysis (3.32 vs. 2.98), while performance on the Quantitative Research module remains relatively close (EduWrite: 2.87 vs. InnoSpark: 2.75). This pattern aligns with our hypothesis that, in vertical domains, data quality density and hierarchically structured training are more critical than scaling parameters alone.

\paragraph{Finding 4: Quantitative research is the most challenging module across all evaluated models.}
Across all model categories, scores on the Quantitative Research module remain lower than those on other modules (e.g., GPT-5: 3.12, versus ≥4.04 on Reviewer/Topic/Policy; EduWrite: 2.87). This consistent difficulty suggests that quantitative scholarly writing requires specialized competencies (e.g., variable operationalization, reliability/validity reasoning, and standards-compliant statistical reporting) that are not sufficiently addressed by general instruction following or enhancements for long-form generation. This observation motivates our subsequent task-level diagnosis and points to a promising direction for targeted improvement.

\paragraph{Finding 5: Reviewing appears easier to saturate than methodology-centered writing.}
Both GPT-5 and Gemini 3 Flash reach 4.03 and 4.07 respectively on the Reviewer module, indicating that current frontier LLMs support language-quality assessment and logic-flow evaluation relatively well. In contrast, modules that require methodological construction and domain-grounded reasoning (e.g., Quantitative and Qualitative Research) remain substantially lower, reinforcing the need for fine-grained evaluation and targeted training strategies.

\section{Related Work \label{relatedwork}}
\subsection{LLM-based Writing Benchmarks}
Benchmarks such as AlpacaFarm~\cite{dubois2023alpacafarm} and MT-Bench~\cite{zheng2023judging} evaluate general-purpose writing via user-like prompts (e.g., emails, blogs) and popularize automated scoring through the LLM-as-a-Judge paradigm. More recent suites like WritingBench~\cite{wu2025writingbench} broaden coverage across writing genres (e.g., creative and functional writing) to provide a more comprehensive profile of general writing ability. Limitation (one sentence): these general writing benchmarks primarily assess readability (fluency and preference) and provide limited measurement of domain depth and methodological rigor, which are essential for academic writing.
To increase domain fidelity, researchers develop vertical benchmarks for generative writing in high-stakes fields. In biomedicine, MultiMedQA~\cite{singhal2023large} and PubMedQA~\cite{jin2019pubmedqa} emphasize factuality and safety in medical QA and literature-oriented generation. In law, LawBench~\cite{fei2024lawbench} and LegalBench~\cite{guha2023legalbench} evaluate legal drafting and case-based reasoning with stronger requirements on logical correctness. Limitation (one sentence): existing vertical writing benchmarks concentrate on biomedical, legal, and natural-science domains and remain sparse for social science—especially education~\cite{bail2024can}—while largely lacking lifecycle-aware, fine-grained evaluation that decomposes research writing into diagnosable modules and atomic tasks.

Building on these insights, EduResearchBench targets educational scholarly writing and introduces the HATD decomposition framework to construct high-quality data and an evaluation framework grounded in the full educational research-and-writing workflow.

\subsection{Benchmarks in Education}
Early education-oriented evaluations largely reuse human standardized tests and measure models’ factual knowledge and comprehension via multiple-choice or short-answer questions~\cite{zhang2023evaluating}. Representative general benchmarks include MMLU~\cite{hendrycksmeasuring} and its Chinese counterpart C-Eval~\cite{huang2023c}, while stage-specific suites such as Gaokao~\cite{zhang2023evaluating} and E-EVAL~\cite{hou2024eval} target particular educational settings (e.g., Chinese college entrance exams and K–12). Subject-focused benchmarks are also common in science and mathematics, including ARC~\cite{clark2018think}, GSM8K~\cite{cobbe2021training}, and Cmath ~\cite{wei2023cmath}. Limitation (one sentence): these exam-style benchmarks primarily assess problem-solving under constrained formats and therefore provide limited coverage of open-ended generation, long-range argumentation, and methodological coherence required in authentic educational research and scholarly writing.
To move beyond factual recall, recent work emphasizes deeper scientific reasoning and complex problem solving. SciBench~\cite{wang2024scibench} evaluates college-level physics, chemistry, and mathematics with multi-step computation and reasoning chains, and SciEval~\cite{sun2024scieval} extends evaluation toward multi-level tasks that approximate research activities across difficulty tiers. Limitation (one sentence): despite stronger reasoning requirements, these benchmarks remain centered on STEM-style “problem-solving” tasks and under-evaluate capabilities critical to social science and education research, such as long-form argumentation, evidence-driven literature synthesis, and qualitative writing.

To address these challenges, we propose EduResearchBench to enable atomic-level, fine-grained evaluation.
\section{Conclusion \label{conclusion}}
This paper targets educational scholarly research writing and introduces EduResearchBench, a comprehensive evaluation platform. To address the limitation that existing evaluations often remain at one-shot end-to-end generation and fail to characterize complex research workflows, we propose HATD, a hierarchical atomic task decomposition framework that structures the full educational research-and-writing process into six specialized research modules and 24 atomic tasks. Building on this decomposition, we construct high-quality data and an automated evaluation pipeline, enabling systematic identification of model capability weaknesses.

Based on over 55K raw academic samples, we curate 11K high-quality instruction pairs to supervisedly fine-tune a specialized model, EduWrite. Experimental results show that EduResearchBench reveals substantial performance differences across research modules; meanwhile, EduWrite leads open-source models and narrows the gap to frontier closed-source systems on multiple core dimensions, further underscoring the importance of high-quality data and structured training strategies in vertical domains.
\bibliographystyle{named}
\bibliography{reference}

\section{Appendix}
\subsection{Prompts}
\subsubsection{Prompt Templates for Data Synthesis and Automated Evaluation}
This appendix reports the prompt templates used in EduResearchBench for (i)data synthesis and (ii)automated evaluation. 
For each atomic task in our HATD taxonomy, we provide bilingual (Chinese/English) prompt templates to ensure cross-lingual reproducibility and consistent task specifications. 
All templates enforce strict output schemas and require models to ground their outputs only in the provided inputs, reducing hallucination and improving the reliability of both synthesized instruction data and evaluation reports.

\subsubsection{Prompt Template for Data Synthesis: Reviewing -- Structure and Logical Flow Analysis}
As an example, we present the bilingual \textbf{data-synthesis} prompt template for the \textit{Reviewing} module, atomic task \textit{Structure and Logical Flow Analysis}. 
Unlike judge prompts that only score model outputs, this template is used to \textbf{generate high-quality instruction--response samples} (i.e., a structured review report) that serve as supervision for SFT and as references for the Standard split.
Accordingly, the prompt (i) assigns the model a strict reviewer role, (ii) requires the analysis to be grounded \textbf{only} in the provided paper text, and (iii) enforces a \textbf{fixed five-part schema} to ensure consistent and reusable outputs:
(1) section identification and core-claim extraction,
(2) argument relationship map,
(3) logic-gap alerts,
(4) section-function evaluation, and
(5) overall evaluation with prioritized revision suggestions.
The input paper is provided via the placeholder \texttt{\{YOUR\_PAPER\_TEXT\}}, and the generated structured report is recorded as the target response.

\begin{table*}[t]
\centering
\small
\renewcommand{\arraystretch}{1.2}
\setlength{\tabcolsep}{7pt}
\caption{Judge = DeepSeek-v3.2. Raw results on \textbf{Chinese (ZH)} data \textbf{before} ZH/EN merging.}
\label{tab:deepseek_zh_raw}
\begin{tabular}{lccccccc}
\toprule
\textbf{Model} & \textbf{Overall} & \textbf{Reviewer} & \textbf{Quant.} & \textbf{Qual.} & \textbf{Topic} & \textbf{Policy} & \textbf{Survey} \\
\midrule
\rowcolor{gray!10} \multicolumn{8}{l}{\textit{Closed-source Models}} \\
GPT-5          & \textbf{3.2996} & \underline{3.5525} & 3.0146 & \textbf{3.5663} & \textbf{3.7923} & 2.9068 & \textbf{2.9650} \\
Gemini 3 Flash & \underline{3.1869} & \textbf{3.6365} & \textbf{3.0348} & \underline{3.5425} & \underline{3.7505} & \underline{2.9575} & \underline{2.6073} \\
\midrule
\rowcolor{gray!10} \multicolumn{8}{l}{\textit{Open-source LLMs}} \\
Qwen-2.5-72B-Instruct & 2.6990 & 3.0184 & 2.6523 & 3.0788 & 3.1074 & 2.4966 & 1.8405 \\
Llama-3-70B-Instruct  & 2.3265 & 2.3464 & 2.5889 & 2.8667 & 2.7702 & 2.1483 & 1.2385 \\
\midrule
\rowcolor{gray!10} \multicolumn{8}{l}{\textit{Capability-enhanced LLMs}} \\
WritingBench (Qwen2.5-7B) & 2.3224 & 2.2865 & 2.3028 & 2.5140 & 2.9365 & 2.4943 & 1.4002 \\
LongWriter (GLM4-9B)      & 2.2845 & 2.4540 & 2.3730 & 2.7665 & 2.5123 & 2.1790 & 1.4224 \\
InnoSpark-72B-1124        & 2.8191 & 2.9560 & \underline{2.7548} & 3.2495 & 3.2580 & 2.7238 & 1.9724 \\
EduWrite (30B)            & 2.9450 & 3.0180 & 2.8616 & 3.3925 & 3.5470 & \textbf{2.7028} & 2.1480 \\
\bottomrule
\end{tabular}

\vspace{3pt}
\begin{minipage}{0.98\linewidth}
\footnotesize
\textit{Note.} \textbf{Bold} indicates the best result and \underline{underlined} indicates the second-best result in each column (computed across all models in this table).
\end{minipage}
\end{table*}

\begin{table*}[t]
\centering
\small
\renewcommand{\arraystretch}{1.2}
\setlength{\tabcolsep}{7pt}
\caption{Judge = DeepSeek-v3.2. Raw results on \textbf{English (EN)} data \textbf{before} ZH/EN merging. (\textit{Reviewer} and \textit{Survey} are not available in EN.)}
\label{tab:deepseek_en_raw}
\begin{tabular}{lccccc}
\toprule
\textbf{Model} & \textbf{Overall} & \textbf{Quant.} & \textbf{Qual.} & \textbf{Topic} & \textbf{Policy} \\
\midrule
\rowcolor{gray!10} \multicolumn{6}{l}{\textit{Closed-source Models}} \\
GPT-5          & \textbf{3.3751} & \textbf{3.0296} & \textbf{3.2835} & \textbf{3.8398} & \textbf{3.3475} \\
Gemini 3 Flash & \underline{2.8946} & \underline{2.5368} & \underline{2.7260} & \underline{3.5260} & \underline{2.7898} \\
\midrule
\rowcolor{gray!10} \multicolumn{6}{l}{\textit{Open-source LLMs}} \\
Qwen-2.5-72B-Instruct & 2.4476 & 2.4379 & 2.2845 & 2.9746 & 2.0935 \\
Llama-3-70B-Instruct  & 2.3026 & 2.3102 & 2.1400 & 2.7122 & 2.0480 \\
\midrule
\rowcolor{gray!10} \multicolumn{6}{l}{\textit{Capability-enhanced LLMs}} \\
WritingBench (Qwen2.5-7B) & 2.2338 & 2.1026 & 1.9430 & 2.8658 & 2.0240 \\
LongWriter (GLM4-9B)      & 2.1370 & 2.1498 & 1.9385 & 2.5213 & 1.9386 \\
InnoSpark-72B-1124        & 2.5629 & 2.5270 & 2.3750 & 3.0465 & 2.3030 \\
EduWrite (30B)            & 2.8315 & 2.6666 & 2.4115 & 3.5363 & 2.7118 \\
\bottomrule
\end{tabular}

\vspace{3pt}
\begin{minipage}{0.98\linewidth}
\footnotesize
\textit{Note.} \textbf{Bold} indicates the best result and \underline{underlined} indicates the second-best result in each column (computed across all models in this table).
\end{minipage}
\end{table*}

\begin{table*}[t]
\centering
\small
\renewcommand{\arraystretch}{1.2}
\setlength{\tabcolsep}{7pt}
\caption{Judge = Doubao-1.6. Raw results on \textbf{Chinese (ZH)} data \textbf{before} ZH/EN merging.}
\label{tab:doubao_zh_raw}
\begin{tabular}{lccccccc}
\toprule
\textbf{Model} & \textbf{Overall} & \textbf{Reviewer} & \textbf{Quant.} & \textbf{Qual.} & \textbf{Topic} & \textbf{Policy} & \textbf{Survey} \\
\midrule
\rowcolor{gray!10} \multicolumn{8}{l}{\textit{Closed-source Models}} \\
GPT-5                & \textbf{4.05} & \underline{4.50} & 3.15 & \underline{4.18} & \textbf{4.36} & 3.83 & \textbf{4.29} \\
Gemini 3 Flash       & \underline{3.80} & \textbf{4.50} & \textbf{3.48} & \textbf{4.25} & \underline{4.28} & \underline{3.89} & \underline{3.80} \\
\midrule
\rowcolor{gray!10} \multicolumn{8}{l}{\textit{Open-source LLMs}} \\
Qwen-2.5-72B-Instruct & 3.49 & 3.79 & \underline{3.18} & 4.15 & 3.73 & 3.10 & 2.99 \\
Llama-3-70B-Instruct  & 2.69 & 2.66 & 2.83 & 3.41 & 3.14 & 2.32 & 1.76 \\
\midrule
\rowcolor{gray!10} \multicolumn{8}{l}{\textit{Capability-enhanced LLMs}} \\
WritingBench (Qwen2.5-7B) & 2.68 & 2.51 & 2.19 & 2.87 & 3.26 & 3.20 & 2.02 \\
LongWriter (GLM4-9B)      & 2.59 & 2.78 & 2.29 & 3.41 & 2.44 & 2.50 & 2.11 \\
InnoSpark-72B-1124        & 3.48 & 3.66 & 2.91 & 3.91 & 3.63 & 3.62 & 3.14 \\
EduWrite (30B)            & 3.75 & 3.65 & 3.08 & 4.06 & 4.19 & \textbf{4.19} & 3.33 \\
\bottomrule
\end{tabular}

\vspace{3pt}
\begin{minipage}{0.98\linewidth}
\footnotesize
\textit{Note.} \textbf{Bold} indicates the best result and \underline{underlined} indicates the second-best result in each column (computed across all models in this table).
\end{minipage}
\end{table*}

\begin{table*}[t]
\centering
\small
\renewcommand{\arraystretch}{1.2}
\setlength{\tabcolsep}{7pt}
\caption{Judge = Doubao-1.6. Raw results on \textbf{English (EN)} data \textbf{before} ZH/EN merging. (\textit{Reviewer} and \textit{Survey} are not available in EN.)}
\label{tab:doubao_en_raw}
\begin{tabular}{lccccc}
\toprule
\textbf{Model} & \textbf{Overall} & \textbf{Quant.} & \textbf{Qual.} & \textbf{Topic} & \textbf{Policy} \\
\midrule
\rowcolor{gray!10} \multicolumn{6}{l}{\textit{Closed-source Models}} \\
GPT-5          & \textbf{3.74} & \textbf{3.29} & \textbf{3.56} & \textbf{3.86} & \textbf{4.25} \\
Gemini 3 Flash & \underline{3.36} & \underline{2.64} & \underline{3.28} & \underline{3.71} & \underline{3.82} \\
\midrule
\rowcolor{gray!10} \multicolumn{6}{l}{\textit{Open-source LLMs}} \\
Qwen-2.5-72B-Instruct & 2.81 & 2.59 & 2.53 & 3.30 & 2.80 \\
Llama-3-70B-Instruct  & 2.51 & 2.31 & 2.28 & 2.61 & 2.83 \\
\midrule
\rowcolor{gray!10} \multicolumn{6}{l}{\textit{Capability-enhanced LLMs}} \\
WritingBench (Qwen2.5-7B) & 2.47 & 2.15 & 2.01 & 2.94 & 2.77 \\
LongWriter (GLM4-9B)      & 2.25 & 2.27 & 2.07 & 2.24 & 2.43 \\
InnoSpark-72B-1124        & 2.88 & 2.79 & 2.73 & 2.73 & 3.28 \\
EduWrite (30B)            & 3.15 & 2.88 & 2.42 & 3.60 & 3.70 \\
\bottomrule
\end{tabular}

\vspace{3pt}
\begin{minipage}{0.98\linewidth}
\footnotesize
\textit{Note.} \textbf{Bold} indicates the best result and \underline{underlined} indicates the second-best result in each column (computed across all models in this table).
\end{minipage}
\end{table*}

\paragraph{Chinese prompt template (data synthesis).}
\begin{ch}{审稿任务-结构与逻辑流分析}{}
请对给定论文进行\textbf{结构与逻辑流分析}。\\
\textbf{要求：}仅按以下 5 个部分输出，不要输出其他内容；分析必须基于论文原文，不杜撰。

\textbf{【一】章节识别与核心论点提取}
\begin{itemize}
  \item 自动识别论文结构（题目/摘要/引言/方法/实验/结果/讨论/结论等）
  \item 对每一章节给出：
  \begin{itemize}
    \item 核心论点/目标
    \item 主要证据或推理链
    \item 与下一章节的过渡是否自然
  \end{itemize}
\end{itemize}

\textbf{【二】论点关系图（文本形式）}
\begin{itemize}
  \item 用节点表示论点：A1, A2, A3...
  \item 用箭头表示支撑：A1 $\rightarrow$ A2（证据：xxx）
  \item 标注弱支撑/缺失支撑关系
\end{itemize}

\textbf{【三】逻辑漏洞提示}
\begin{itemize}
  \item 指出论证不足、矛盾、概念跳跃、不当推理
  \item 每条包含：位置 + 原因 + 修补建议
\end{itemize}

\textbf{【四】章节功能评估}
\begin{itemize}
  \item 评估：引言/方法/结果/讨论/结论是否分别完成对应功能
  \item 每章给出：符合 / 部分符合 / 不符合 + 理由
\end{itemize}

\textbf{【五】总体评价}
\begin{itemize}
  \item 逻辑连贯性评分：优秀 / 良好 / 一般 / 较差（含理由）
  \item 最优先修改的 3 条建议
\end{itemize}

\textbf{输入：论文全文}
\par\noindent\textbf{$\langle\langle\langle$论文开始$\rangle\rangle\rangle$}\\
\{YOUR\_PAPER\_TEXT\}\\
\textbf{$\langle\langle\langle$论文结束$\rangle\rangle\rangle$}

\end{ch}

\paragraph{English prompt template (data synthesis).}
\begin{prm}{Analysis of Structure and Logical Flow}{}
\textbf{\# Role}\\
You are a strict reviewer of logic and structure in education research papers. \ldots

\textbf{\# Task}\\
Conduct a ``Structure and Logic Flow Analysis'' of the paper below. Output \textbf{strictly} in the specified format. \ldots

\textbf{[1] Section Identification and Core Claim Extraction}
\begin{itemize}
  \item Identify paper structure (title, abstract, introduction, methods, \ldots).
  \item For each section, output:
  \begin{itemize}
    \item Core claim/goal \ldots
    \item Main evidence or reasoning chain \ldots
    \item Transition role to the next section (whether natural) \ldots
  \end{itemize}
\end{itemize}

\textbf{[2] Argument Relationship Map (text format)}
\begin{itemize}
  \item Claims as nodes: A1, A2, A3, \ldots
  \item Support as arrows: A1 $\rightarrow$ A2 (evidence: \ldots)
  \item Mark weak/missing support \ldots
\end{itemize}

\textbf{[3] Logic Gap Alerts}
\begin{itemize}
  \item Identify issues: insufficient argumentation, contradictions, conceptual leaps, \ldots
  \item Each item includes: location + reason + fix \ldots
\end{itemize}

\textbf{[4] Section Function Evaluation}
\begin{itemize}
  \item Introduction: problem and research gap \ldots
  \item Methods: reproducibility \ldots
  \item Results: data-only reporting \ldots
  \item Discussion: interpretation and contributions \ldots
  \item Conclusion: answers research question \ldots
\end{itemize}
For each section: ``Meets / Partially Meets / Does Not Meet'' + reasons \ldots

\textbf{[5] Overall Evaluation}
\begin{itemize}
  \item Logical coherence rating: Excellent / Good / Fair / Poor \ldots
  \item Top 3 highest-priority revision suggestions \ldots
\end{itemize}

\textbf{Input: Full paper text}
\par\noindent\textbf{$\langle\langle\langle$Paper Begin$\rangle\rangle\rangle$}\\
\{YOUR\_PAPER\_TEXT\}\\
\textbf{$\langle\langle\langle$Paper End$\rangle\rangle\rangle$}

\end{prm}

\subsubsection{Prompt Template for Automated Evaluation (LLM-as-a-Judge): Reviewing -- Structure and Logical Flow Analysis}
\label{app:judge_prompt_reviewer_logic}

We also provide the bilingual \textbf{LLM-as-a-Judge} prompt template for the same atomic task, which is used during evaluation to assess model-generated outputs.
To improve robustness and explainability, the judge is instructed to produce \textbf{(i) a critique} followed by \textbf{(ii) a scalar score} on a 1--5 scale, while adhering to the same five-part schema for diagnostic comparability.
The evaluated content is injected through placeholders such as \texttt{\{YOUR\_PAPER\_TEXT\}} and \texttt{\{MODEL\_OUTPUT\}}.

\paragraph{Chinese prompt template (judge).}
\begin{ch}{评测提示词-结构与逻辑流分析}{}
你是一名高级教育学学术审稿与元评价专家。你的任务不是评价论文，而是评价“论文结构与逻辑流分析报告”的质量。\\
你必须基于论文原文与分析报告作出判断，\textbf{不得臆测或杜撰}。\ldots

\textbf{【任务背景（必须参考）】}
\begin{itemize}
  \item 分析报告目标：深层语义分析；识别各章节核心论点与证据；判断论点逻辑关系与过渡；发现逻辑漏洞；评估章节是否支撑研究问题。\ldots
  \item 正确输出应包含：逻辑连贯性分析；论点关系图/支撑结构；逻辑漏洞提示（位置/原因/缺失证据）；章节功能评估。\ldots
  \item 关键能力：NLU；语义推理；结构化表达；逻辑评估。\ldots
\end{itemize}

\textbf{【评分维度（1--5 分）】}\\
请基于原文与分析报告，对每个维度给出：\textbf{分数 + 评价 + 证据}（不得只给分数）。
\begin{enumerate}
  \item 章节识别与论点提取准确性：结构划分是否正确；核心论点与证据是否准确；是否误判/遗漏。\ldots
  \item 论点关系与逻辑链分析质量：是否建立清晰支撑链（如 A$\rightarrow$B$\rightarrow$C）；过渡/跳跃/缺环判断是否基于事实；关系图是否具体。\ldots
  \item 逻辑漏洞发现与解释深度：是否指出无依据结论/矛盾/跳跃；是否定位明确并解释原因；是否避免捏造。\ldots
  \item 章节功能评估是否合理：是否与原文一致；是否判断各章是否支撑研究问题；是否避免过度主观。\ldots
  \item 全面性、深度与可解释性：是否覆盖核心内容；结论是否清晰结构化可读；是否避免空洞评价。\ldots
\end{enumerate}

\textbf{【总评分标准】}
\begin{itemize}
  \item 最终评分为 1--5 分（整数），依据五维度总分映射：\ldots
  \item 总分区间 $\rightarrow$ 总评分映射：\ldots
\end{itemize}

\textbf{【必须输出】}
\begin{enumerate}
  \item 对分析报告质量的详细评价 \ldots
  \item 分维度评论（每个维度单独评价与理由）\ldots
  \item 末行仅输出最终评分（1--5 的一个数字）\ldots
\end{enumerate}

\textbf{输入：论文原文}
\par\noindent\textbf{$\langle\langle\langle$论文开始$\rangle\rangle\rangle$}\\
\{论文原文\}\\
\textbf{$\langle\langle\langle$论文结束$\rangle\rangle\rangle$}

\textbf{输入：待评分的分析报告}
\par\noindent\textbf{$\langle\langle\langle$分析报告开始$\rangle\rangle\rangle$}\\
\{分析报告\}\\
\textbf{$\langle\langle\langle$分析报告结束$\rangle\rangle\rangle$}

\end{ch}

\paragraph{English prompt template (judge).}
\begin{prm}{Structure and Logical Flow Analysis}{}

You are a senior peer-review and meta-evaluation expert in education research. Your task is not to evaluate the paper itself, but to evaluate the quality of the ``Structure and Logic Flow Analysis Report.''\\
You must judge \textbf{based on the paper text and the report only}; do \textbf{not} speculate or fabricate.\ldots

\textbf{[Task Background (must reference)]}
\begin{itemize}
  \item The report's goals: deep semantic analysis; identify each section's core claims and supporting evidence; assess argument relations and transitions; check logic gaps/leaps/unsupported conclusions; assess whether each section supports the research question/objectives.\ldots
  \item A correct report should include: coherence analysis; an argument map/support structure; logic-gap alerts (location/reason/missing evidence); section-function evaluation.\ldots
  \item Required capabilities: NLU; semantic reasoning; structured presentation; logical evaluation.\ldots
\end{itemize}

\textbf{[Scoring Dimensions (1-5)]}\\
For \textbf{each} dimension, provide: \textbf{score + commentary + evidence} (do not give score only).
\begin{enumerate}
  \item Accuracy of section identification and claim extraction: correct structure; accurate core claims/evidence; no major missing/misclassified sections.\ldots
  \item Quality of argument relations and logic-chain analysis: clear support chains (e.g., A$\rightarrow$B$\rightarrow$C); transitions/leaps/gaps judged from text; map is concrete rather than vague.\ldots
  \item Depth of logic-gap detection and explanation: identifies unsupported conclusions/contradictions/leaps; precise localization and reasons; avoids fabricated issues.\ldots
  \item Reasonableness of section-function evaluation: consistent with the paper; assesses whether sections support the research question; avoids overly subjective claims.\ldots
  \item Coverage, depth, and explainability: covers core content; conclusions are clear/structured/readable; avoids hollow judgments.\ldots
\end{enumerate}

\textbf{[Overall Score Rule]}
\begin{itemize}
  \item Output a final integer score from 1 to 5, based on the mapped sum of the five dimensions:\ldots
  \item Mapping from total (5-25) to final (1-5): \ldots
\end{itemize}

\textbf{[Required Output]}
\begin{enumerate}
  \item Detailed evaluation of the report quality \ldots
  \item Per-dimension comments (each dimension separately) \ldots
  \item Final line: output \textbf{only} one digit (1--5) as the final score \ldots
\end{enumerate}

\textbf{Input: Full Paper Text}
\par\noindent\textbf{$\langle\langle\langle$Paper Begin$\rangle\rangle\rangle$}\\
\{FULL\_PAPER\_TEXT\}\\
\textbf{$\langle\langle\langle$Paper End$\rangle\rangle\rangle$}

\textbf{Input: Analysis Report to Score}
\par\noindent\textbf{$\langle\langle\langle$Report Begin$\rangle\rangle\rangle$}\\
\{ANALYSIS\_REPORT\}\\
\textbf{$\langle\langle\langle$Report End$\rangle\rangle\rangle$}

\end{prm}

\subsection{Experiment Results}
Tables~\ref{tab:doubao_zh_raw} and~\ref{tab:doubao_en_raw} report the \textbf{raw} agent-level results under the \textbf{Doubao-1.6} judge \emph{before} merging Chinese (ZH) and English (EN) evaluations.
The ZH set covers six agents (\textit{Reviewer, Quantitative, Qualitative, Topic, Policy, Survey}), whereas the EN set does \textbf{not} include \textit{Reviewer} and \textit{Survey}, and therefore only reports four agents (\textit{Quantitative, Qualitative, Topic, Policy}).
All scores are averaged over the corresponding tasks for each agent.

Tables~\ref{tab:deepseek_zh_raw} and~\ref{tab:deepseek_en_raw} report the \textbf{raw} agent-level results under the \textbf{DeepSeek-v3.2} judge \emph{before} merging Chinese (ZH) and English (EN) evaluations.
The ZH set covers six agents (\textit{Reviewer, Quantitative, Qualitative, Topic, Policy, Survey}), whereas the EN set does \textbf{not} include \textit{Reviewer} and \textit{Survey}, and therefore only reports four agents (\textit{Quantitative, Qualitative, Topic, Policy}).
For \textit{Gemini 3 Flash} on ZH, the original table did not provide an \textit{Overall} score; we compute it by task-count weighting over the six agents (2/5/4/4/4/5, summing to 24).

\subsection{Case Study}
This case study focuses on the Hotspot Extraction task within the Topic Recommendation module. The task requires identifying key educational "hotspots" from the provided academic text corpus, aligning them with current educational trends, policy relevance, and future development directions.

The input corpus discusses various aspects of Latin sentence translation and teaching methods, highlighting complex teaching approaches and educational reforms. The goal of the hotspot extraction task is to identify content that is high in frequency, relevant to future educational directions (such as AI and digitalization), and closely tied to policy reforms.

\paragraph{Key Findings.}
\textbf{EduWrite} stands out for its ability to synthesize and present the most relevant educational hotspots clearly and precisely. It fully adheres to the output format and employs professional, domain-specific terminology, making its responses the most accurate and well-structured.
\textbf{GPT-5} provides solid output but lacks depth in policy relevance and identification of specific trends, resulting in a slightly lower score compared to EduWrite.
\textbf{Gemini 3 Flash} offers a broader synthesis but fails to fully capture the key trends in the input, lacking the necessary depth and policy relevance.
\textbf{Qwen2.5-72B-Instruct} performs well but includes extraneous details and policy implications not present in the input corpus, which diminishes its overall effectiveness.

\begin{case}{Case}{Hotspot Extraction in Topic Recommendation}
"The application of linguistic theories to the problem of teaching pupils to translate complex Latin sentences into English... Teachers who use the traditional grammar/translation method seem much more definite about methods of teaching Latin grammar than they are about methods of teaching translation... The introduction of stave analysis as a method of teaching translation from Latin into English is not a confession that the main purpose of teaching Latin is to teach translation rather than reading."
\textit{... and so on.}
\end{case}

\begin{case}{Model Responses and Evaluation}{Model Responses and Judge Scores}
\textbf{GPT-5 Response:}

\textit{Strengths:} GPT-5 provides a reasonable set of educational themes, including "integration of linguistic theory and pedagogy" and "scientific teaching methods." The response adheres to the format well but does not fully capture the relevance of the educational trends mentioned in the corpus. While accurate in terms of output, GPT-5 lacks depth in policy context, failing to draw connections between the identified hotspots and current educational policies.

\textit{Evaluation:} GPT-5 receives a moderate score (4.0) for its response. It adheres to the format but lacks precision in identifying the most relevant hotspots, particularly in terms of policy implications.

\textbf{Gemini 3 Flash Response:}

\textit{Strengths:} Gemini 3 Flash provides a more extensive synthesis of educational concepts but falls short in terms of specificity. The response touches on key terms like "language teaching" and "literacy" but misses finer details that align with the educational trends or reforms indicated in the input corpus.

\textit{Evaluation:} Gemini 3 Flash scores moderately (3.5). It provides a broad overview but misses out on critical trends, such as the role of modern linguistic theories in educational practices. It also lacks focus on policy relevance, which is crucial for the hotspot extraction task.

\textbf{EduWrite Response:}

\textit{Strengths:} EduWrite excels in this task by accurately extracting relevant education hotspots, such as "stave analysis as a visual aid" and "dependency markers in Latin education," both of which are highly relevant to the input. EduWrite synthesizes the core educational trends in a precise and structured manner, using academic terminology such as "dependency parsing" and "string analysis." The response is clear, concise, and highly aligned with the input corpus.

\textit{Evaluation:} EduWrite achieves the highest score (5.0) for its response. It demonstrates exceptional precision in identifying the relevant hotspots and uses professional academic language, ensuring clarity and relevance. Its adherence to format and depth of content make it stand out from the other models.

\textbf{Qwen2.5-72B-Instruct Response:}

\textit{Strengths:} Qwen2.5-72B-Instruct identifies some educational themes but includes certain policy implications that were not explicitly mentioned in the input corpus. It also includes some less relevant details, making the response somewhat verbose. While it does capture some key trends, it lacks the level of precision and relevance seen in EduWrite’s output.

\textit{Evaluation:} Qwen2.5-72B-Instruct scores 3.8. It offers a functional analysis but includes extraneous details and policy implications not covered in the input. Its verbosity detracts from the conciseness and relevance of the output.

\end{case}

\end{document}